\documentclass[conference]{IEEEtran}
\usepackage{times}

\usepackage{multicol}
\usepackage[bookmarks=true]{hyperref}
\usepackage[numbers,sort]{natbib}
\usepackage{times}
\usepackage{wrapfig}
\usepackage{xspace}
\usepackage{graphicx}
\usepackage[version=latest]{pgf}
\usepackage{tabularx}
\usepackage{amsmath}
\usepackage{amssymb}
\usepackage{url}
\usepackage{bm}
\usepackage{color}
\usepackage{colortbl}
\usepackage{subfigure}
\usepackage{multirow}
\usepackage{calrsfs}
\usepackage{graphicx,import}
\DeclareGraphicsExtensions{.pdf,.jpeg,.png}
\graphicspath{{figures/}}
\usepackage{balance}

\usepackage{etoolbox}
\makeatletter
\patchcmd{\@makecaption}
  {\scshape}
  {}
  {}
  {}
\makeatletter
\patchcmd{\@makecaption}
  {\\}
  {.\ }
  {}
  {}
\makeatother

\newcommand{\vt}{\mathbf{t}}
\newcommand{\mt}{\mathbf{T}}

\newcommand{\dotsangle}{13}
\newcommand{\dotsxshift}{.2ex}
\newcommand{\dotsyshift}{0.2ex}
\newcommand{\rdots}{\hspace{\dotsxshift}\raisebox{\dotsyshift}{\rotatebox{\dotsangle}{$\ddots$}}}

\newcommand{\tstart}{0}
\newcommand{\tend}{T}

\DeclareMathOperator*{\argmin}{arg\,min}

\newcommand{\bx}{\mathbf{x}}

\newcommand{\bu}{\mathbf{u}}
\newcommand{\bz}{\mathbf{z}}
\newcommand{\bn}{\mathbf{n}}
\newcommand{\bp}{\mathbf{p}}

\newcommand{\bq}{\mathbf{q}}

\newcommand{\bb}{\mathbf{b}}
\newcommand{\bv}{\mathbf{v}}
\newcommand{\bC}{\mathbf{C}}

\newcommand{\figref}[1]{Fig.~\ref{fig:#1}}
\newcommand{\secref}[1]{Sec.~\ref{sec:#1}}
\newcommand{\equref}[1]{Eq.~\ref{eq:#1}}

\newcommand{\tabref}[1]{Tab.~\ref{tab:#1}}
\DeclareMathAlphabet{\pazocal}{OMS}{zplm}{m}{n}
\definecolor{jpgreen}{rgb}{0.0 0.7 0.3}

\newcommand{\vc}{\mathbf{c}}
\newcommand{\vy}{\mathbf{y}}

\begin{document}

% paper title
\title{Observability-Aware Trajectory Optimization for Self-Calibration with Application to UAVs}

% You will get a Paper-ID when submitting a pdf file to the conference system
%\author{Author Names Omitted for Anonymous Review. Paper-ID 99}

\author{\authorblockN{Karol Hausman${^*}$, James Preiss${^*}$, Gaurav S. Sukhatme${^*}$}
\authorblockA{${^*}$Department of Computer Science\\
University of Southern California\\
Los Angeles, CA 90089, USA\\
Email: hausman, japreiss, gaurav@usc.edu}
\and
\authorblockN{Stephan Weiss${^\dagger}$}
\authorblockA{${^\dagger}$Institute of Smart System Technologies\\
Alpen-Adria-Universitat Klagenfurt\\
Klagenfurt, 9020, Austria\\
Email: stephan.weiss@aau.at}}

\maketitle

\begin{abstract}
We study the nonlinear observability of a system's states in view of \emph{how well} they are observable and what control inputs would improve the convergence of their estimates.
We use these insights to develop an observability-aware trajectory-optimization framework for nonlinear systems that produces trajectories well suited for self-calibration. 
Common trajectory-planning algorithms tend to generate motions that lead to an unobservable subspace of the system state, causing suboptimal state estimation. 
We address this problem with a method that reasons about the quality of observability while respecting system dynamics and motion constraints to yield the optimal trajectory for rapid convergence of the self-calibration states (or other user-chosen states). 
Experiments performed on a simulated quadrotor system with a GPS-IMU sensor suite demonstrate the benefits of the optimized observability-aware trajectories when compared to a covariance-based approach and multiple heuristic approaches. 
Our method is $\thicksim$80x faster than the covariance-based approach and achieves better results than any other approach in the self-calibration task. We applied our method to a waypoint navigation task and achieved a $\thicksim$2x improvement in the integrated RMSE of the global position estimates and $\thicksim$4x improvement in the integrated RMSE of the GPS-IMU transformation estimates compared to a minimal-energy trajectory planner.
\end{abstract}

\IEEEpeerreviewmaketitle

\section{Introduction}
\label{sec:intro}

State estimation is a core capability for autonomous robots. 
For any system, it is desirable to estimate the state at any point in time as accurately as possible.
Accurate state estimation is crucial for robust control strategies and serves as the foundation for higher-level planning and perception.
In addition to the states directly used for system control, such as position, velocity, and attitude, more recent work also estimates internal states that calibrate the sensor suite of the system~\cite{ROB:ROB21466}. 
These so-called \emph{self-calibration states} include all information needed to calibrate the sensors against each other -- such as the position and attitude of one sensor with respect to another -- as well as their intrinsic parameters such as measurement bias.

In general, these states could be estimated \emph{a priori} using offline calibration techniques. 
The advantages of including self-calibration states in the \emph{online} state estimator are threefold: i) the same implementation of the self-calibrating estimator can be used for different vehicles ii) the state estimator can compensate for errors in the initialization or after collisions and iii) the platform does not require any offline calibration routine because it can self-calibrate itself while operating.

\begin{figure}
  \centering
  \includegraphics[width=0.99\columnwidth]{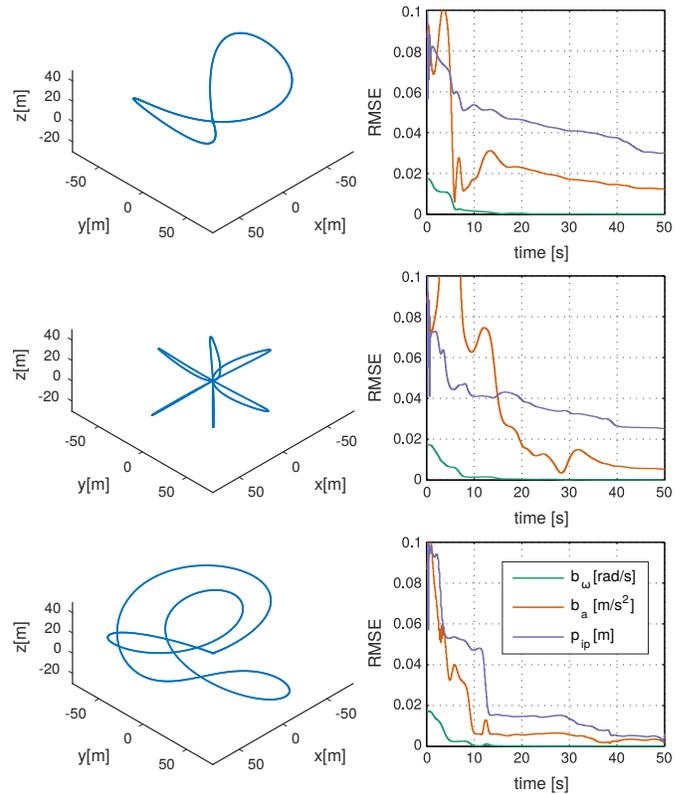}
  \caption{Root Mean Squared Error (RMSE) convergence of EKF self-calibration (accelerometer and gyroscope biases $\bb_a$, $\bb_\omega$ and position of the GPS sensor $\bp^p_i$) state estimates for a) figure-eight trajectory, b) star trajectory, c) optimal trajectory from our method. We introduced additional yaw motion for a) and b) trajectories in order to improve state estimation of these heuristics.}
  \label{fig:cover}
\end{figure}

Including self-calibration states in the estimator has many advantages, but it comes at an important cost:  the dimensionality of the state vector increases while the number of measurements remains unchanged. 
This cost often leads to the requirement of engineered system inputs to render all states observable, i.e. the system needs a tailored trajectory that might require extra time or energy compared to an observability-unaware trajectory.
This requirement holds true not only for the initial self-calibration but also throughout the mission.
Usually, the self-calibration motion is executed by an expert operator who controls the vehicle and continuously checks if the states have converged to reasonable values. 
During the mission, the expert operator must take care to excite the system sufficiently to keep all states observable.
More importantly, in autonomous missions, trajectory-planning algorithms that minimize energy use 
may generate trajectories that lead to an unobservable subspace of the system state.

In this paper, we present a framework that optimizes trajectories for self-calibration.
The resulting trajectories avoid unobservable subspaces of the system state during the mission.
We develop a cost function that explicitly addresses the quality of observability of system states.
Our method takes into account motion constraints and yields an optimal trajectory for fast convergence of the self-calibration states or any other user-chosen states. 
The presented theory applies to any (non)linear system and it is not specific to a particular realization of the state estimator such as KF, EKF, or UKF. 
While past approaches have focused on analyzing the environment to compute \emph{where} to move to obtain informative measurements for state estimation~\cite{bry11icra}, we assume the presence of accurate measurements\footnote{Advanced sensors and state estimators are nowadays able to obtain accurate measurements in a large variety of different environments~\cite{Li2013b}} and focus on \emph{how} to move to generate motions that render the full state space observable.

In order to evaluate our method, we conduct several experiments using a simulated Unmanned Aerial Vehicle (UAV) with GPS-IMU sensor suite, as this is a common scenario that illuminates the problem of self-calibration.
An example of the performance of the self-calibration framework is presented in~\figref{cover}, where an optimized trajectory outperforms common calibration heuristics usually chosen by experts in terms of speed and accuracy of the state convergence.

The key contributions of our approach are:
\begin{itemize}
    \item we present a method that is able to predict the quality of state estimation based on the vehicle's ego-motion rather than on the perceived environment; the method takes into account system dynamics, measurements and the nonlinear observability analysis.
    \item Our method is carried out on the nonlinear continuous system without making any state-estimator-specific assumptions.
    \item We demonstrate a full self-calibration-based trajectory optimization framework that is readily adjustable for any dynamical system and any set of states of the system.
    \item We show that the observability-aware trajectory optimization can be also used for the waypoint navigation task which results in more accurate state estimation.
\end{itemize}

\section{Related Work}
\label{sec:rw}

Previous work on improving state estimation of a system has mainly focused on analyzing the environment to choose informative measurements~\cite{bryson04acra,julian2012distributed}. 
With the arrival of robust visual-inertial navigation solutions (e.g. Google Tango\footnote{\url{https://www.google.com/atap/project-tango/}}), sufficiently accurate measurements can be obtained most of the time during a mission without special path reasoning. 
These results give us the opportunity to shift focus towards other aspects of the trajectory, in particular, its suitability for self-calibration. 
In~\cite{Maye06102015}, the authors find the best set of measurements from a given trajectory to calibrate the system. Unobservable parameters are locked until the trajectory has sufficient information to make them observable. 
The analysis is performed on the linearized system and analyzes a given trajectory rather than generating an optimal convergence trajectory.
\citet{s150203154} analyzed a specific marine system and developed a trajectory to calibrate it based on heuristics. 
The approach is not generally applicable to other systems. 
Other approaches analyze the final covariance of the system when simulating it on a test trajectory: 
\citet{4177715} maximize the inverse of the covariance at the final time step and use this cost in an optimization procedure. 
Similarly, in~\cite{achtelik13icra} the authors sample a subset of the state space with a Rapidly Exploring Random Tree approach~\cite{bry11icra} and optimize for the final covariance of the system.
These approaches are sample-based techniques discretizing their environment and state space. 
The discretization and linearization steps induce additional errors and may lead to wrong results similar to the well-known rank issue when analyzing a system in its linearized instead of the nonlinear form~\cite{Hesch13112013}.

From the nonlinear observability analysis described in~\cite{hermann1977nonlinear}, \citet{krener2009measures} develop a measure of observability rather than only extracting binary information on the observability of a state. 
\citet{6580494} make use of this measure to generate trajectories that optimize the convergence of states that are directly visible in the sensor model.
We make use of this definition and extend the approach to analyze the quality of observability of states that are not directly visible in the sensor model. 
This way, we can also generate motions leading to trajectories that optimize the convergence of, e.g., IMU biases.

Trajectory optimization has been successfully used in many different applications including robust perching for fixed-wing vehicles~\cite{moore2014robust}, locomotion for humanoids~\cite{kuindersma2015optimization} and manipulation tasks for robotic arms~\cite{levine2015learning}.
Since we evaluate our system using a model of an Unmanned Aerial Vehicle (UAV), we present the related work on the trajectory optimization in this area.
\citet{mellinger2011minimum} use trajectory optimization of differentially flat variables of a quadrotor to obtain minimum-snap trajectories.
~\citet{richter2013polynomial} presented an extension of this approach to generate fast quadrotor paths in cluttered environments using an unconstrained QP.
In~\cite{mueller14iros}, the authors generate risk-aware trajectories using the formulation developed by~\citet{van2011lqg} with a goal of safe quadrotor landing.
\citet{hausman2015cooperative} use the framework of trajectory optimization to generate controls for multiple quadrotors to track a mobile target.
Finally,~\citet{moore2014robust} use LQR-trees to optimize for a trajectory that leads to robust perching for a fixed-wing vehicle.
In this work, we use trajectory optimization to generate paths for self-calibration, which are evaluated on a simulated quadrotor system.
Our work augments other trajectory-optimization-based approaches by providing an observability-aware cost function that can be used in combination with other optimization objectives.

\section{Problem Formulation and Fundamentals}

\subsection{Motion and Sensor Models}
 
We assume the following nonlinear system dynamics, i.e. the motion model:
\begin{align*}
    \dot{\bx} &= f(\bx,\bu,\mathbf{\delta}),
    %\label{eq:system1}
\end{align*}
where $\bx$ is the state, $\bu$ are the control inputs and $\mathbf{\delta}$ is noise caused by non-perfect actuators and modelling errors.

For sensory output we use the nonlinear sensor model:
\begin{align*}
    \bz &= h(\bx, \mathbf{\epsilon}),
    %\label{eq:system2}
\end{align*}
where $\bz$ is the sensor reading and $\mathbf{\epsilon}$ is the sensor noise.

It is often the case that there are certain elements in the state vector $\bx$, such as sensor biases, that stay constant according to the motion model, i.e. their values are independent of the controls $\bu$ and other states $\bx$. 
In this paper, we will call these entities self-calibration states $\bx_{sc}$. 
%\begin{align}
%    \dot{\bx}_{sc} &= f(\mathbf{\delta}).
%\end{align}

\subsection{Nonlinear Observability Analysis}
\label{sec:nonlinear-observability-analysis}

The observability of a system is defined as the possibility to compute the initial state of the system given the sequence of its inputs $\bu(t)$ and measurements $\bz(t)$.
A system is said to be \emph{globally observable} if there exist no two points $\bx_0(0)$, $\bx_1(0)$ in the state space with the same input-output $\bu(t)$-$\bz(t)$ maps for any control inputs.
A system is \emph{weakly locally observable} if there is no point $\bx_1(0)$ with the same input-output map in a neighborhood of $\bx_0(0)$ for a specific control input.

Observability of linear as well as nonlinear systems can be determined by performing a \emph{rank test} where the system is observable if the rank of the observability matrix is equal to the number of states.
In the case of a nonlinear system, the nonlinear observability matrix is constructed using the Lie derivatives of the sensor model $h$.
Lie derivatives are defined recursively with zero-noise assumption. The 0-th Lie derivative is the sensor model itself, i.e.:
\begin{align*}
    L_0^h = h(\bx),
\end{align*}
the next Lie derivative is constructed as:
\begin{align*}
    L_{i+1}^h = \frac{\partial}{\partial t}{L}_{i}^h = \frac{\partial {L}_{i}^h}{\partial \mathbf{x}}\frac{\partial \mathbf{x}}{\partial t} = \frac{\partial {L}_{i}^h}{\partial \mathbf{x}}f(\mathbf{x},\mathbf{u}).
\end{align*}
One can observe that Lie derivatives with respect to the sensor model are equivalent to the respective time derivatives of the sensory output $\bz$:
\begin{align*}
    \dot{\bz} = \frac{\partial}{\partial t}\bz(t) = \frac{\partial}{\partial t}h(\mathbf{x}(t)) = \frac{\partial h}{\partial \mathbf{x}}\frac{\partial \mathbf{x}}{\partial t} = \frac{\partial h}{\partial \mathbf{x}}f(\mathbf{x},\mathbf{u}) = {L}_{1}^h.
\end{align*}
By continuing to compute the respective Lie derivatives one can form the matrix:
\begin{align*}
O(\bx,\bu) &=
\begin{bmatrix}
\nabla L_0^h&\nabla {L}_{1}^h&\nabla {L}_{2}^h&\dots
\end{bmatrix}^T,
\end{align*}
where $\nabla L_0^h = \frac{\partial L_0^h}{\partial \bx}$ and $\dot{\bz}=\frac{\partial \bz}{\partial t}$.

The matrix $O(\bx,\bu)$ formed from the sensor model and its Lie derivatives is the nonlinear observability matrix.
Following~\citet{hermann1977nonlinear}, if the observability matrix has full column rank, then the state of the nonlinear system is weakly locally observable.
Unlike linear systems, nonlinear observability is a local property that is input-- and state-dependent.

It is worth noting that the observability of the system is a binary property and does not quantify how \emph{well} observable the system is. 
This limits its utility for gradient-based methods.
We address this issue in the next section.

\section{Quality of Observability}
\label{sec:quality}

Following~\citet{krener2009measures} and according to the definition presented in~\secref{nonlinear-observability-analysis}, we introduce the notion of quality of observability.

A state is said to be \emph{well} observable if the system output changes significantly when the state is marginally perturbed~\cite{WeissPhD2012}.
A state with this property is robust to measurement noise and it is highly distinguishable within some proximity where this property holds. 
Conversely, a state that leads to a small change in the output, even though the state value was extensively perturbed, is defined as \emph{poorly} observable. 
In the limit, the measurement does not change even if we move the state value through its full range. 
In this case, the state is unobservable~\cite{hermann1977nonlinear}.

\subsection{Taylor Expansion of the Sensor Model}
In order to model the variation of the output in relation to a perturbation of the state, we approximate the sensor model using the $n$-th order Taylor expansion about a point $t_0$:

\begin{align*}
 %&h_{t_0}(\bx(t),\bu(t)) = h^0(\bx(t_0),\bu(t_0)) + (t-t_0)h^1(\bx(t_0),\bu(t_0))\\
 %&+ \ldots + h^n(\bx(t_0),\bu(t_0))\frac{(t-t_0)^n}{n!}
 h_{t_0}(\bx(t),\bu(t)) &= \sum_{i=0}^n \frac{(t-t_0)^i}{i!} h^i(\bx(t_0),\bu(t_0)),
\end{align*}
where $h_{t_0}$ represents the Taylor expansion of $h$ about $t_0$ with the following Taylor coefficients $h^0, h^1, \ldots, h^n$:
\begin{align*}
h^0(\bx(t_0),\bu(t_0)) &= h(\bx(t_0),\bu(t_0)) = L^h_0(\bx(t_0),\bu(t_0)) \\
h^1(\bx(t_0),\bu(t_0)) & = \frac{\partial}{\partial t} (h(\bx(t_0),\bu(t_0))) = L^h_1(\bx(t_0),\bu(t_0)) \\
%h^i(\bx(t_0),\bu(t_0)) & = L^h_i(\bx(t_0),\bu(t_0)). 
& \vdots \\
h^n(\bx(t_0),\bu(t_0)) & = L^h_n(\bx(t_0),\bu(t_0)). 
\end{align*}

%By introducing $\delta{t} = t-t_0$ and $h(\bx(t),\bu(t)) = h(t)$ the resulting Taylor expansion of the sensor model about $t_0$ is equal to:
%\begin{align}
%h_{t_0}(t) &= L^h_0(t_0) + \delta{t}L^h_1(t_0) + \frac{\delta{t}^2}{2!}L^h_2(t_0)\\
%&+ \dots + \frac{\delta{t}^n}{n!}L^h_n(t_0).     
%\end{align}

Using this result, one can also approximate the state derivative of the sensor model $\frac{\partial}{\partial \bx}h(\bx(t),\bu(t))$.
For brevity, we introduce the notation $\delta{t} = t-t_0$, $h_{t_0}(t) = h_{t_0}(\bx(t),\bu(t))$, and we omit the arguments of the Lie derivatives: 
\begin{align*}
\frac{\partial}{\partial \bx}h_{t_0}(t) &= \sum_{i=0}^n 
%\frac{\partial}{\partial \bx}L^h_0 + \delta{t}\frac{\partial}{\partial \bx}L^h_1 + \dots + \frac{\delta{t}^n}{n!}\frac{\partial}{\partial \bx}L^h_n\\
 \frac{\delta{t}^i}{i!}\nabla L^h_i.
%\label{eq:taylor-derivative}
\end{align*}
%Using the property of the Lie derivatives
%\begin{align}
%    L^h_1 = \nabla L^h_0 \dot{\bx}
%\end{align}
%where $\nabla = \frac{\partial}{\partial \bx}$,
%one may reformulate~\equref{taylor-derivative} as following
%\begin{align}
%\frac{\partial}{\partial t}h(\bx(t))|_{t=0} &= \nabla L^h_0 \dot{\bx} + t\nabla L^h_1 \dot{\bx} + \frac{t^2}{2}\nabla L^h_2 \dot{\bx} + \dots + \frac{t^n}{n!}\nabla L^h_n \dot{\bx} \\
%\frac{\partial}{\partial t}h(\bx(t))|_{t=0} &= (\nabla L^h_0 + t\nabla L^h_1 + \frac{t^2}{2}\nabla L^h_2 + \dots + \frac{t^n}{n!}\nabla L^h_n) \dot{\bx}. 
%\end{align}
%By multiplying both sides by $\delta t$ and one obtains:
%\begin{align}
%\delta{h}(\bx(t))|_{t=0} &= (\nabla L^h_0 + t\nabla L^h_1 + \frac{t^2}{2}\nabla L^h_2 + \dots + \frac{t^n}{n!}\nabla L^h_n) \delta \bx,
%\end{align}
This result in matrix form is:
\begin{align}
%\frac{\partial}{\partial \bx}h_{t_0}(t) &=  \begin{bmatrix}1 & \delta{t} & \frac{\delta{t}^2}{2} & \dots & \frac{\delta{t}^n}{n!})\end{bmatrix} \begin{bmatrix} \nabla L^h_0\\\nabla L^h_1\\\nabla L^h_2\\ \vdots\\ \nabla L^h_n \end{bmatrix}  \\
\frac{\partial}{\partial \bx}h_{t_0}(t) &=  \begin{bmatrix}I & \delta{t}I & \frac{\delta{t}^2}{2}I & \dots & \frac{\delta{t}^n}{n!}I\end{bmatrix} O(\bx(t),\bu(t)),
\label{eq:taylor-final}
\end{align}
where $O(\bx(t),\bu(t))$ is the nonlinear observability matrix.

\equref{taylor-final} describes the Jacobian of the sensor model $h$ with respect to the state $\bx$ around the time $t_0$. 
Using this Jacobian, we are able to predict the change of the measurement with respect to a small perturbation of the state. 
This prediction not only incorporates the sensor model but it also implicitly models the dynamics of the system via high order Lie derivatives. 
Hence, in addition to showing the effect of the states that directly influence the measurement, \equref{taylor-final} also reveals the effects of the varying control inputs and the states that are not included in the sensor model.
This will prove useful in~\secref{measure}.

\subsection{Observability Gramian}
%observability gramian
In addition to the change in the output with respect to the state perturbation, one needs to take into account the fact that different states can have different influence on the output.
Thus, a large effect on the output caused by a small change in one state can swamp a similar effect on the output caused by a different state and therefore, \emph{weaken} its observability. 
In order to model these interactions, following~\cite{krener2009measures}, we employ the local observability Gramian:
\begin{align}
     W_o(\tstart,\tend) &= \int_{\tstart}^{\tend} \Phi(\tstart, t)^T H(t)^T H(t) \Phi(\tstart, t) dt,
     \label{eq:gramian}
\end{align}
where $\Phi(\tstart, t)$ is the state transition matrix (see~\cite{krener2009measures} for details), $H(t)$ is the Jacobian of the sensor model $H(t) = \frac{\partial}{\partial \bx}h(t)$ and the trajectory spans the time interval $t \in [0, T]$.

Since a nonlinear system can be approximated by a linear time-varying system by linearizing its dynamics about the current trajectory, one can also use the local observability Gramian for nonlinear observability analysis.
If the rank of the local observability Gramian is equal to the number of states, the original nonlinear system is locally weakly observable~\cite{hermann1977nonlinear}.

\citet{krener2009measures} introduced measures of observability that are based on the condition number or the smallest singular value of the local observability Gramian.
Unfortunately, the local observability Gramian is difficult to compute for any nonlinear system. 
In fact, it can only be computed in closed form for certain simple nonlinear systems. 
In order to solve this problem, the local observability Gramian can be approximated numerically by simulating the sensor model for small state perturbations, resulting in the \emph{empirical local observability Gramian}~\cite{krener2009measures}:
\begin{align}
    W_o \approx \frac{1}{4\epsilon^2} \int_{\tstart}^{\tend} \begin{bmatrix} \Delta{\bz_1^T(t)}\\ \vdots \\\Delta{\bz_n^T(t)} \end{bmatrix} [ \Delta{\bz_1(t)} \, \dots \, \Delta{\bz_n(t)}] dt,
    \label{eq:gramian-approx}
\end{align}
where $\Delta{\bz_i} = \bz^{+i} - \bz^{-i}$ and $\bz^{\pm i}$ is the simulated measurement when the state $\bx^i$ is perturbed by a small value $\pm \epsilon$.
The empirical local observability Gramian in~\equref{gramian-approx} converges to the local observability Gramian in~\equref{gramian} for $\epsilon \rightarrow 0$.

The main disadvantage of this numerical approximation is that it cannot approximate the local observability Gramian for the states that do not appear in the sensor model.
As $\epsilon \rightarrow 0$ this approximation replaces the state transition matrix $\Phi(\tstart, t)$ with the identity matrix.
This relieves the burden of finding an analytical solution for $\Phi(\tstart, t)$, however, it also eliminates any effects on the local observability Gramian caused by the states that are not in the sensor model.
Thus, it becomes difficult to reason about the observability of these states using this approximation.
We address this problem in the following section.

\subsection{Measure of Observability}
\label{sec:measure}
In order to present the hereby proposed measure of observability concisely, we introduce the following notation:
\begin{align*}
    K_{t_0} (t) &= \frac{\partial}{\partial \bx}h_{t_0}(t) = \frac{\partial}{\partial \bx}h_{t_0}(\bx(t), \bu(t)).
\end{align*}

Following the definition of the local observability Gramian, we use the Taylor expansion of the sensor model to approximate the local observability Gramian:
\begin{align}
%    W_o(\tstart,\tend) &= \int_{\tstart}^{\tend} \Phi(\tstart, t)^T H(t)^T H(t) \Phi(\tstart, t) dt\\
    W_o(\tstart,\tend, \Delta{t}) &\approx \int_{\tstart}^{\tend}  K_t (t + \Delta{t})^T K_t (t + \Delta{t}) dt,
    \label{eq:gramian-final}
\end{align}
where $\Delta{t}$ is a fixed horizon that enables us to see the effects of the system dynamics. 
In order to measure the quality of observability we use the smallest singular value of the approximated local observability Gramian $W_o(\tstart,\tend, \Delta{t})$.

In contrast to the empirical local observability Gramian, our formulation is able to capture input-output dependencies that are not visible in the sensor model.
We achieve this property by incorporating higher order Lie derivatives that are included in the observability matrix. 
Intuitively, at each time step, we use the Taylor expansion of the sensor model about the current time step $t$ to approximate the Jacobian of the measurement in a fixed time horizon $\Delta{t}$.
We use this approximation to estimate the local observability Gramian which is integrated over the entire trajectory.

In order to measure the observability of a subset of the states, one can use the smallest singular value of the submatrix of the local observability Gramian that includes only the states of interest. 
We use this property to focus on different self-calibration states of the system.

\section{Trajectory Representation and Optimization}
\label{sec:trajectory}

\subsection{Differentially Flat Trajectories}
In order to efficiently represent trajectories, we consider differentially flat systems~\cite{van1997real}. 
A system is differentially flat if all of its inputs $\bx, \bu$ can be represented as a function of flat outputs $\mathbf{y}$ and their finite derivatives $\mathbf{\tilde{y}}$, i.e.:
\begin{align*}
    \bx &= \mathbf{\zeta}(\mathbf{y},\dot{\mathbf{y}}, \ddot{\mathbf{y}}, ..., \overset{(n)}{\mathbf{y}}) = \mathbf{\zeta}(\mathbf{\tilde{y}})\\
    \bu &= \mathbf{\psi}(\mathbf{y},\dot{\mathbf{y}}, \ddot{\mathbf{y}}, ..., \overset{(m)}{\mathbf{y}}) = \mathbf{\psi}(\mathbf{\tilde{y}}).
\end{align*}
For the rest of this paper, we express robot trajectories as flat outputs because this is the minimal representation that enables us to deduce the state and controls of the system over time.

\subsection{Constrained Trajectory Representation using Piecewise Polynomials}
\label{sec:constrained}
Similar to~\citet{mueller14iros}, we represent a trajectory by a 
$k$-dimensional, $d$-degree piecewise polynomial, composed of $q$ pieces:
\begin{align*}
\vy(t)=
\begin{cases}
	P_1 \vt(t) &\text{if  } t_0 \le t < t_1 \\
	\vdots\\
	P_q \vt(t) &\text{if  } t_{q-1} \le t \le t_q,
\end{cases}
\end{align*}
%\begin{align}
    %\mathbf{y}(t) = \sum_{i=0}^d \bp_i t^i = P \mathbf{t}(t)
%\end{align}
where $P_i$ is the $k \times (d+1)$ matrix of polynomial coefficients for the $i$th polynomial piece,
and $\vt$ is the time vector, i.e.:
\begin{align*}
	\vt(t) = \begin{bmatrix}t^0 & t^1 & \dots & t^d\end{bmatrix}^T.
    %P = \begin{bmatrix} \bp_0 & \bp_1 & \dots & \bp_d \end{bmatrix}, \mathbf{t}(t) = \begin{bmatrix}t^0 & t^1 & \dots & t^d\end{bmatrix}^T
\end{align*}

We formulate constraints on the initial and final positions and derivatives of a trajectory
as a system of linear equations~\cite{mueller14iros}. For example:
\begin{align*}
    \vc_1 &= \vy(\tstart) = P_1 \vt(\tstart) \\
    \vc_2 &= \dot{\vy}(\tstart) = P_1 \dot{\vt}(\tstart) \\ 
	\vdots \\
    \vc_f &= \overset{(n)}{\vy}(\tend) = P_q \overset{(n)}{\vt}(\tend),
\end{align*}
where $\vc_1, \vc_2, ..., \vc_f$ are the trajectory constraints and $\dot{\vt}$ is the trivial derivative $\dot{\vy}(t) = P_i\dot{\vt}(t)$.
In matrix form, the initial constraints appear as:
\begin{align}
\label{eq:constraints}
    \begin{bmatrix} \mathbf{c}_1 & \mathbf{c}_2 & \dots \end{bmatrix} = P_1 \begin{bmatrix} \mathbf{t}(\tstart) & \mathbf{\dot{t}}(\tstart) & \dots \end{bmatrix}
\end{align}
and the final constraints are defined simlarly.

In addition to start and end constraints, a physically plausible trajectory must be continuous up to the $\beta$-th derivative:
\begin{align*}
    P_1 \vt(t_1) = P_2 \vt(t_1) &\: \hdots\:  P_1 \overset{(\beta)}{\vt}(t_1) = P_2 \overset{(\beta)}{\vt}(t_1)\\ 
    & \enspace \, \vdots  \\
    P_{q-1} \vt(t_{q-1}) = P_q \vt(t_{q-1}) &\: \hdots\:  P_{q-1} \overset{(\beta)}{\vt}(t_{q-1}) = P_q \overset{(\beta)}{\vt}(t_{q-1}).
\end{align*}
To compactly express the evaluation of a polynomial and its first $\beta$ derivatives at a point in time, we define the time matrix:
\begin{align*}
\mt_i = \begin{bmatrix} \vt(t_i) & \dot{\vt}(t_i) & \dots & \overset{(\beta)}{\vt}(t_i) \end{bmatrix}.
\end{align*}
We may thus express the smoothness constraints as a banded linear system:
\begin{align}
\label{eq:ppconstraints}
\begin{bmatrix} P_1 & \dots & P_q \end{bmatrix}
\begin{bmatrix} 
 \mt_1 &        &            &            \\
-\mt_1 & \mt_2  &            &            \\
       & \rdots & \rdots     &            \\
       &        & -\mt_{q-2} &  \mt_{q-1} \\
       &        &            & -\mt_{q-1} \\
\end{bmatrix} = 0.
\end{align}
If we add equations in the form of \equref{constraints} for the initial and final constraints, the resulting linear system completely expresses the problem constraints.

With an appropriately high polynomial degree $d$, \equref{ppconstraints} forms an underdetermined system.
Therefore, we can use the left null space of the constraint matrix as the optimization space. 
Any linear combination of the left null space may be added to the particular solution to form a different polynomial that still satisfies the trajectory constraints.
For the particular solution, we use the minimum-norm solution from the Moore-Penrose pseudoinverse, which gives smooth trajectories.

The described change of variables significantly reduces the dimensionality of the optimization problem and eliminates all equality constraints.
The only remaining constraints are nonlinear inequalities related to the physical limits of the motor torques, etc., required to execute the path.

\subsection{Numerical Stability for Constrained Trajectory}
The linear system presented in the previous section tends to be ill-conditioned. 
Its condition number grows exponentially with the polynomial degree~\cite{pan2015bad} and it is exacerbated by longer time intervals in the polynomial pieces.
Using the formulation from~\cite{mellinger2011minimum}, we scale the time duration of the problem such that each polynomial piece lies in a time interval of $\leq 1$ second.
This produces a better-conditioned matrix whose solution can easily be converted back into a solution to the original problem.

\subsection{The Optimization Objective}
\label{sec:objective}
The goal of this work is to find a trajectory that will provide an optimal convergence of the self-calibration parameters of a nonlinear system. 
In order to achieve this goal, we aim at minimizing the cost function of the following form:
\begin{align*}
    %\int_{\tstart}^{\tend} o(\mathbf{\tilde{y}}(t))dt %+ cQ(\mathbf{\tilde{y}}(t))
    \argmin_{\tilde{\mathbf{y}}(\tstart), ..., \tilde{\mathbf{y}}(\tend)} o(\mathbf{\tilde{y}}(t)),
\end{align*}
where $o(\mathbf{\tilde{y}}(t))$ is the observability-dependent cost that is directly related to the convergence of the self-calibration states in the estimator.

In case of the hereby presented measure of quality of observability, $o(\mathbf{\tilde{y}}(t))$ is: 
\begin{align*}
o(\mathbf{\tilde{y}}(t)) = \sigma_{min}(W_o(\tstart,\tend, \Delta{t})),
\end{align*}
where $\sigma_{min}(W_o(\tstart,\tend, \Delta{t}))$ is the minimum singular value of the approximated local observability Gramian $W_o$ described in~\equref{gramian-final}. 

To the best of our knowledge, the only other cost function that reflects the convergence of the states of the system is based on the EKF covariance.
Minimizing the trace of the covariance results in minimizing the uncertainty about the state for all of its individual dimensions~\cite{beinhofer2013robust} and yields better results than optimizing its determinant (i.e. mutual information)~\cite{hausman2015cooperative}.
Therefore, we employ the covariance-trace cost function that integrates the traces of the covariance submatrices that are responsible for the self-calibration states.
We use this method as one of the baselines for our approach.

As described in~\secref{constrained}, introducing the new constrained trajectory representation enables us to pose trajectory optimization as an unconstrained optimization problem and reduce its dimensionality.
However, in order to ensure the physical plausibility of the trajectory, we still need to optimize it subject to physical limits of the system.
We represent the physical inequality constraints as nonlinear functions of the differentially flat variables.

For optimization we use the implementation of the Sequential Quadratic Programming (SQP) method with nonlinear inequality constraints from Matlab Optimization Toolbox.

\section{Example Application to UAVs with IMU-GPS State Estimator}
\label{sec:application}

We demonstrate the presented theory on a simulated quadrotor with a 3-DoF position sensor (e.g. GPS) and a 6-DoF inertial measurement unit (IMU).
This is a simple, widely popular sensor suite, but it presents a challenging self-calibration task, as there is limited intuition for what kind of trajectory would make the states well observable.  
Although we present experiments for the quadrotor, we emphasize that the presented theory can be applied to a variety of nonlinear systems.

\subsection{EKF for IMU-GPS Sensor Suite}
As a realization of the state estimator of the quadrotor, we employ the popular Extended Kalman Filter (EKF).
The EKF continuously estimates state values by linearizing the motion and sensor model around the current mean of the filter.
It recursively fuses all controls $\bu_{1:k}$ and sensor readings $\bz_{1:k}$ up to time $k$
and maintains the state posterior probability: 
\begin{align*}
  p(\mathbf{x}_k \mid \mathbf{z}_{1:k}, \mathbf{u}_{1:k}) = \mathcal{N}(\hat{\mathbf{x}}_k, \Sigma_k)
\end{align*}
as a Gaussian with mean $\hat{\mathbf{x}}_k$ and covariance $\Sigma_k$.
In particular, we use the indirect formulation of an iterated EKF~\cite{lynen2013robust} where the state prediction is driven by IMU measurements. 
We choose this state estimator due its ability to work with various sensor suites and proven robustness in the quadrotor scenario.

The state consists of the following:
\begin{align}
  \bx^T &= [{\bp^{i}_w}^T, {\bv^{i}_w}^T, {\bq^{i}_w}^T, {\bb_\omega}^T, {\bb_a}^T, {\bp^{p}_i}^T], 
\label{eq:state_core}
\end{align}
where $\bp^{i}_w$, $\bv^{i}_w$ and $\bq^{i}_w$ are the position, velocity and orientation (represented as a quaternion) of the IMU in the world frame,
$\bb_w$ and $\bb_a$ are the gyroscope and accelerometer biases,
and ${\bp^{p}_i}^T$ is is the relative position between the GPS module and the IMU.

The state is governed by the following differential equations:
\begin{align}
\dot{\bp}^{i}_w  &=  \bv^{i}_w \nonumber  \\
\dot{\bv}^{i}_w  &=  \bC_{(\bq^{i}_w)}^T (\mathbf{a}_m - \bb_a - \bn_a) - \mathbf{g} \nonumber \\
\dot{\bq}^{i}_w  &=  \frac{1}{2} \Omega (\mathbf{\omega}_m - \bb_\mathbf{\omega} - \bn_\omega)  \bq^{i}_w \nonumber \\ 
%\label{eq:system_model}\\
 \dot{\bb_w} &= \bn_{\bb_\mathbf{\omega}}, \dot{\bb_a} = \bn_{\bb_a}, 
 \label{eq:random-process}
\end{align}
where $\bC_{(\bq)}$ is the rotation matrix obtained from the quaternion $\bq$, $\Omega(\mathbf{\omega})$ is the quaternion multiplication matrix of $\mathbf{\omega}$,  $\mathbf{a}_m$ is the measured acceleration, and $\mathbf{\omega}_m$ is the  angular velocity with white Gaussian noise $\bn_a$ and $\bn_\omega$. 
Since the IMU biases can change over time, they are modeled as random processes %(see~\equref{random-process})
%\begin{align}
%  \dot{\bb_\omega} = \bn_{\bb_\omega}, \quad \dot{\bb_a} = \bn_{\bb_a}
%\end{align}
where $\bn_{\bb_w}$ and $\bn_{\bb_a}$ are assumed to be zero-mean Gaussian random variables.

Starting from the initial state defined in \equref{state_core}, we define the error state as:
\begin{align*}
      \tilde{\bx}^T &= [\Delta{\bp^i_w}^T, \Delta{\bv^{i}_w}^T, {\delta\Theta^{i}_w}^T, \Delta{\bb_\omega}^T, \Delta{\bb_a}^T, \Delta{\bp^{p}_i}^T],
%\label{eq:state_error}
\end{align*}
where $\tilde{\bx}$ is the error between the real state value $\bx$ and the state estimate $\hat{\bx}$. For quaternions the error state is defined as: $\delta\bq = \bq \otimes \hat{\bq} \approx {[1 \quad \frac{1}{2}\delta\Theta^T]}^T$.

In this setup, the self-calibration error states $\tilde{\bx}_{sc}$ are the gyroscope and accelerometer biases $\bb_\omega, \bb_a$ and position of the GPS sensor in the IMU frame $\bp^{p}_i$:
\begin{align*}
    \tilde{\bx}^T_{sc} &= [\Delta{\bb_\omega}^T, \Delta{\bb_a}^T, \Delta{\bp^{p}_i}^T].
\end{align*}

Using the IMU-GPS state vector in~\equref{state_core}, the system dynamics in~\equref{random-process}, and assuming the connection between the IMU and the GPS sensor is rigid, we define the GPS sensor model as:
\begin{align*}
    \bz_{gps} &=  h(\bx, \bn_{\bz_{gps}}) = \bp_w^i + \bC_{(\bq^{i}_w)}^T \bp_i^p + \bn_{\bz_{gps}},
\end{align*}
where $\bn_{\bz_{gps}}$ is white Gaussian measurement noise.

The nonlinear observability analysis in~\cite{kelly2011visual} and~\cite{WeissPhD2012} shows that the system is fully observable with appropriate inputs.

\subsection{Differentially Flat Outputs and Physical Constraints of the System}

As shown by~\citet{mellinger2011minimum} the quadrotor dynamics are differentially flat. 
This means that a quadrotor can execute any smooth trajectory in the space of flat outputs as long as the trajectory respects the physical limitations of the system.
The flat outputs are $x,y,z$ position and yaw $\theta$:
\begin{align*}
    \mathbf{y} = [x,y,z,\theta]^T.
\end{align*}
The remaining extrinsic states, i.e. roll and pitch angles, are functions of the flat outputs and their derivatives.
In order to ensure that trajectories are physically plausible we place inequality constraints on 3 entities: the thrust-to-weight ratio ($\leq 1.5$), angular velocity ($\leq \pi \frac{rad}{s}$), and angular acceleration ($\leq 5\pi\frac{rad}{s^2}$).
These values are rough estimates for a small-size quadrotor.

\section{Experimental Results} 
\label{sec:experiments}

\subsection{Experimental Setup}
We evaluate the proposed method in simulation using the quadrotor described in~\secref{application}.
The simulation environment enables extensive testing with ground truth self-calibration states that would not be possible for a real robot.
We represent trajectories as degree-6 piecewise polynomials with continuity up to the 4th derivative.
In all experiments, we require trajectories with zero velocity and acceleration at the beginning and end points.

The quadrotor has a GPS sensor that is positioned $\bp^p_i = [0.1\,\,0.1\,\,0.1]^T$m away from the IMU and produces measurements with standard deviation of 0.2m.
The accelerometer and the gyroscope have initial biases of $\bb_a = [0.05\,\,0.05\,\,0.05 ]^T$ m/s$^2$ and $\bb_\omega = [0.01\,\,0.01\,\,0.01]^T$ rad/s respectively.
These are common values for real quadrotor systems that we have used.
The initial belief is that all the self-calibration states are zero.
Thus, a bad self-calibration trajectory will fail to converge the state estimate of the system.

\subsection{Evaluation of Various Self-Calibration Routines}

\begin{figure}
  \centering
  \includegraphics[width=0.99\columnwidth]{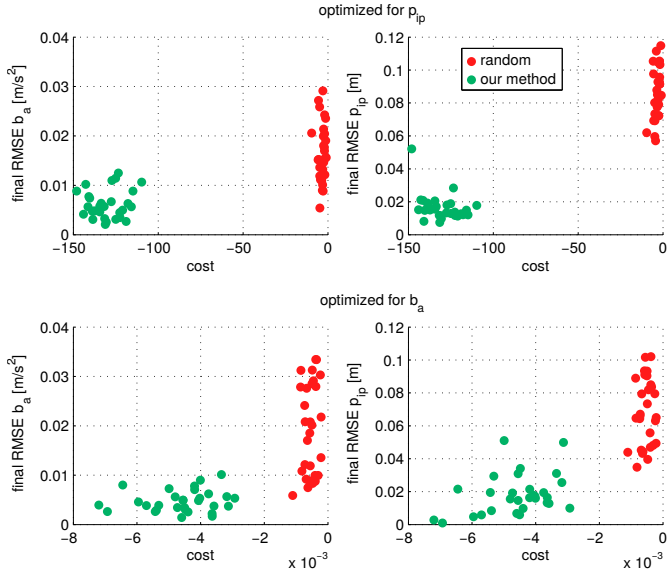}
  \caption{Self-calibration task: results obtained for optimizing for different objectives. Top row: random and optimized results when optimizing for $\bp^p_i$. Bottom row: random and optimized results when optimizing for $\bb_a$. The left and right column show the final RMSE for $\bb_a$ and $\bp^p_i$ respectively.}
  \label{fig:ba-vs-pip}
\end{figure}

\begin{figure}
  \centering
  \includegraphics[width=0.99\columnwidth]{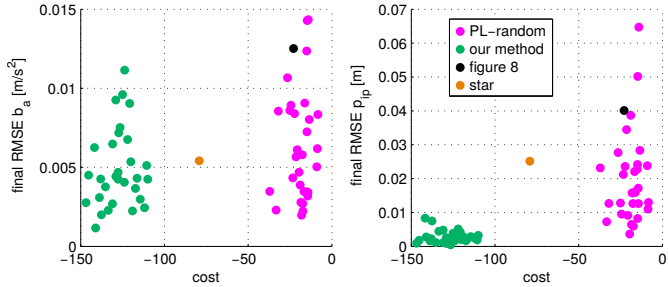}
  \caption{Self-calibration task: final RMSE values for the accelerometer bias $\bb_a$ and the GPS position in the IMU frame $\bp^p_i$ obtained using optimization (green) and 3 different heuristics: star and figure eight trajectories from~\figref{cover} and randomly sampled trajectories that are close of the physical limits of the system.}
  \label{fig:scatter-all}
\end{figure}

To evaluate the influence of choosing different states to construct the local observability Gramian, we compared two optimization objectives: i) the local observability Gramian constructed using the position states with the $\bp^p_i$ states, and ii) the position states with the $\bb_a$ states.
Initial tests showed that $\bb_\omega$ converges quickly for almost any trajectory, so we did not include it in the evaluation.
For the self-calibration task we require trajectories to start and end at the same position. 
We generated random trajectories by sampling a zero-mean Gaussian distribution for each optimization variable, i.e. each component of the left null space of the piecewise polynomial constraint matrix (\equref{ppconstraints}).
We then used each random trajectory as an initial condition for nonlinear optimization to produce an optimized trajectory.

\figref{ba-vs-pip} shows the optimization results using both objectives.
The optimized trajectories significantly outperformed the randomly generated ones.
The two self-calibration states $\bp^p_i$ and $\bb_a$ are co-related in our system, i.e. optimizing for one state also leads to improved performance on the other state. 
However, one can observe that the trajectories optimized for $\bb_a$ yield improved $\bb_a$ final RMSE values compared to those of trajectories optimized for $\bp_i^p$, and analogously
the trajectories optimized for $\bp^p_i$ yield better $\bp_i^p$ results than those optimized for $\bb_a$.   
Due to the small differences between results in the case of the accelerometer bias $\bb_a$ and the larger difference for the position of the GPS sensor $\bp^p_i$, we chose to conduct further experiments using the $\bp^p_i$ objective.

\figref{scatter-all} shows results from the same experiment for a number of differently constructed trajectories.
\emph{PL-random} is a more competitive set of random trajectories generated by choosing larger random null space polynomial weights and discarding trajectories that violated the quadcopter's physical limits.
The remaining trajectories are therefore likely to contain velocities and accelerations that are near the physical limits, which should lead to better observability.
\emph{Figure 8} and \emph{star} are the heuristic trajectories presented in~\figref{cover},
and \emph{our method} are trajectories generated from our optimization framework using the \emph{PL-random} trajectories as initial conditions.
While the \emph{star} trajectory and some of the \emph{PL-random} trajectories perform well on $\bb_a$, our approach outperforms all other methods on $\bp_i^p$.

\begin{figure}
  \centering
  \includegraphics[width=0.99\columnwidth]{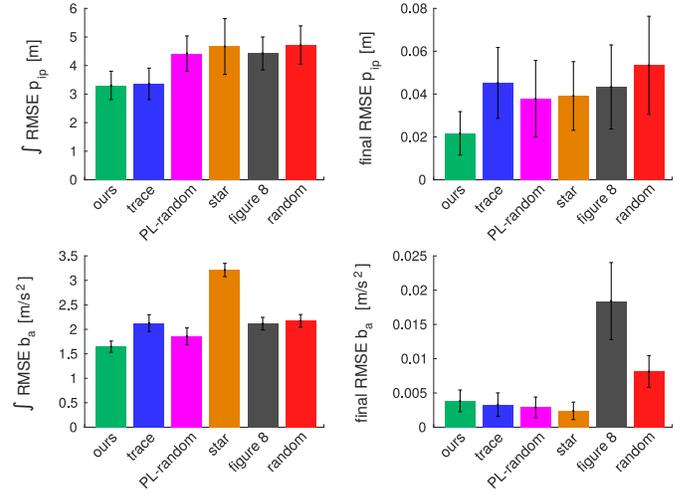}
  \caption{Self-calibration task: statistics collected over 50 runs of the quadrotor EKF using 6 different trajectories: ours - optimized trajectory using the hereby defined observability cost; trace - optimized trajectory using the covariance-trace cost function; PL-random - randomly sampled trajectory that is very close to the physical limits of the system; star, figure 8 - heuristics-based trajectories presented in~\figref{cover}; random - randomly sampled trajectory that satisfies the constraints. Top left: GPS position integrated RMSE, top right: GPS position final RMSE, bottom left: accelerometer bias integrated RMSE, bottom right: accelerometer bias final RMSE.}
  \label{fig:bars}
\end{figure}

In order to more extensively test the different self-calibration strategies, we collected statistics over 50 EKF simulations for a single representative trajectory from each strategy. 
\figref{bars} summarizes our results in terms of the RMSE integrated over the entire trajectory and the final RMSE for accelerometer bias $\bb_a$ and GPS position $\bp^p_i$.
Results show that our approach outperforms all baseline approaches in terms of the final and integrated RMSE of the GPS position $\bp^p_i$.
The only method that is able to achieve a similar integrated RMSE value for the GPS position is the covariance-trace-based optimization described in~\secref{objective}.
However, it takes approximately 13 hours for the optimizer to find that solution, versus approximately 10 minutes with our method.
The main reason for this is that in order to estimate the trace of the covariance of the EKF, one needs to perform matrix inversion at every time step,
which is more computationally expensive than the integration of the local observability Gramian and the one singular value decomposition used by our approach. 
The integrated RMSE of the accelerometer bias $\bb_a$ also suggests that our approach is able to make this state converge faster than in other methods.
Nevertheless, a few other trajectories such as covariance-trace-based and PL-random were able to perform well in this test.
This is also visible in the case of the final RMSE of the accelerometer bias $\bb_a$ where the first four methods yield similar results. 
While our method is slightly worse than the covariance-trace-based and the two heuristics-based approaches,
given the standard deviation of the measurement (0.2m) and the final RMSE values of the bias being below 0.005 $\frac{m}{s^2}$ (which is less than 10\% of the initial bias RMSE), we consider the trajectories from all 4 methods to have converged this estimate.

\subsection{Evaluation of an Example Waypoint Trajectory}
In addition to the self-calibration task, we applied our method to a waypoint navigation task.
With minor extensions, the piecewise polynomial constraint matrix formulation in~\equref{ppconstraints} can satisfy position and derivative constraints along the path in addition to start and endpoint constraints.
We compare a trajectory optimized using our method to a minimum snap trajectory computed using the method from~\cite{mellinger2011minimum}.

\begin{figure}
  \centering
  \includegraphics[width=0.99\columnwidth]{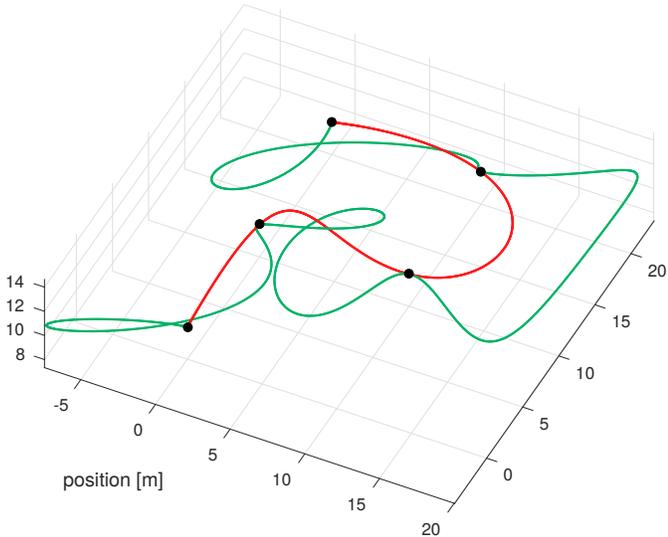}
  \caption{Waypoint navigation task: minimum snap trajectory (red) vs. optimized trajectory using our method (green). Position waypoints are shown in black. 
  }  
  \label{fig:waypoint}
\end{figure}

\figref{waypoint} shows both of the optimized trajectories. 
The trajectory optimized using our method is much more complex than a simple min-snap trajectory because it aims to yield well-observable states.
The results in~\tabref{waypoints} show that our trajectory yields 4x better GPS sensor position estimates and 2x better position estimates than the min-snap trajectory. 
We note that even though the observability-aware trajectory is longer and more complex, which makes the state estimation harder, the resulting estimates are still significantly better than the min-snap trajectory.
This result supports the intuition that sensor calibration can have significant influence on the estimation of other system states.

\balance

\subsection{Discussion}
The presented results indicate that our approach is able to outperform other baselines at the task of estimating the position of the GPS sensor in the IMU frame.
However, it yields comparable results with other methods regarding the accelerometer bias.
Even though for this simple system one can think of heuristics that performs reasonably well, these may not generalize for more complex systems.

The main advantage of the presented framework lies in its generality -- it is applicable to any nonlinear system -- and the fact that it can be combined with other objectives as long as they can be represented in the cost function.

The waypoint navigation task presents another useful application of our technique, and shows that the influence of the GPS-IMU position estimate on the quality of position estimation can be significant (see~\tabref{waypoints}).

\begin{table}
\begin{center}
  \begin{tabular}{| l || c | c |}
    \hline
     & min-snap & our method \\ \hline \hline
    $\bp^i_w$ $\int$ RMSE & $6.06 \pm 1 m$ & $3.69 \pm 0.31 m$ \\ \hline
    $\bp^p_i$ $\int$ RMSE & $3.40 \pm 0.83 m$ & $0.85 \pm 0.22 m$ \\
    \hline
  \end{tabular}
  \caption {Waypoint navigation task: statistics of the integrated RMSE values for position $\bp^i_w$ and GPS sensor position in the IMU frame $\bp^p_i$ collected over 50 runs. Both trajectories take 50 seconds.}
  \label{tab:waypoints}
\end{center}
\end{table}

\section{Conclusion} 
\label{sec:conclusion}

We introduced an observability-aware trajectory optimization framework that is applicable to any nonlinear system and produces trajectories that are well suited for self-calibration.
In contrast to existing approaches, our method moves the focus from \emph{where} to go during a mission to \emph{how} to achieve the goal while staying well-observable.
The presented results performed for a simulated quadrotor system with a GPS-IMU sensor suite demonstrate the benefits of the optimized observability-aware trajectories compared to other heuristics and a covariance-based approach.
For the self-calibration task we were able to achieve almost 2x better final RMSE values for the GPS-IMU position state than all the baseline approaches and comparable converged values for the accelerometer and gyroscope biases.
Our method runs $\thicksim$80x faster than the only other generic baseline approach that is applicable to other systems, and it achieved better results.

The presented method was also applied to a waypoint navigation task and achieved almost 2x better integrated RMSE of the position estimate and more than 4x better integrated RMSE of the GPS-IMU position estimate than the minimum snap trajectory.

In the future, we plan to test this method on multi-sensor fusion systems where the observability of the states is of even greater importance and the self-calibration states have bigger influence on the other states.
The next steps also include evaluating the optimized trajectories on real UAVs and other robotic systems.

%\newpage

\bibliographystyle{plainnat}
\bibliography{references}

\end{document}